\documentclass[runningheads]{llncs}

\usepackage{times}
\usepackage{url}
\usepackage{latexsym}
\usepackage{amsmath}
\usepackage{comments}
\usepackage{csquotes}
\usepackage{subcaption}
\newcommand{\repeatthanks}{\textsuperscript{\thefootnote}}
\usepackage[symbol]{footmisc}

\newauthor[SH]{Simon}{pink}   
\newauthor[LP]{Lidia}{magenta}     
\newauthor[Ez]{Elaine}{green}  
\newauthor[MT]{Mikko}{yellow}  
\newauthor[JM]{Jani}{cyan}     

\title{Topic modelling discourse dynamics in historical newspapers}

\renewcommand*{\thefootnote}{\fnsymbol{footnote}} 

\author{Jani Marjanen\inst{1}\thanks{Equal contribution.}\orcidID{0000-0002-3085-4862} \and 
Elaine Zosa\inst{2}\repeatthanks\orcidID{0000-0003-2482-0663} \and \\
Simon Hengchen\inst{3}\orcidID{0000-0002-8453-7221} \and
Lidia Pivovarova\inst{2}\orcidID{0000-0002-0026-9902} \and \\
Mikko Tolonen\inst{1}\orcidID{0000-0003-2892-8911}}
\authorrunning{J. Marjanen et al.}
\institute{
Helsinki Computational History Group, University of Helsinki \and
Department of Computer Science, University of Helsinki \and
Språkbanken, University of Gothenburg\footnote{SH was affiliated with the University of Helsinki for most of this work.}
 \\
\email{firstname.lastname@\{helsinki.fi,gu.se\}}}

\date{}

\begin{document}
\maketitle
\begin{abstract}
  This paper addresses methodological issues in diachronic data analysis for historical research. We apply two families of topic models (LDA and DTM) on a relatively large set of historical newspapers, with the aim of capturing and understanding discourse dynamics. Our case study focuses on newspapers and periodicals published in Finland between 1854 and 1917, but our method can easily be transposed to any diachronic data. Our main contributions are a) a combined sampling, training and inference procedure for applying topic models to huge and imbalanced diachronic text collections; b) a discussion on the differences between two topic models for this type of data; c) quantifying topic prominence for a period and thus a generalization of document-wise topic assignment to a discourse level; and d) a discussion of the role of humanistic interpretation with regard to analysing discourse dynamics through topic models.
\end{abstract}

\renewcommand*{\thefootnote}{\arabic{footnote}}  
\setcounter{footnote}{0}

\section{Introduction}

This paper reports our experience on studying discursive change in Finnish newspapers from the second half of the nineteenth century. We are interested in grasping broad societal topics, discourses that cannot be reduced to mere words, isolated events or particular people. Our long-lasting goal is to investigate a global change in the presence of such topics and especially finding discourses that have disappeared or declined and thus could easily slip away in modern research. We believe that these research questions are better approached in a data-driven way without deciding what we are looking for beforehand, though the choice of the most suitable techniques for such research is still an open problem. 

In this paper we focus on developing methodology. Choosing available algorithms for analysis guides possible outcomes as they are designed to be operationalised in certain ways. Approaching our goal with mere word counts is counterproductive due to the sparseness of the language and the variety of discourse realisations in a given text. Further, word counts are unreliable with historical data due to never ending language change, spelling variations and text recognition errors. 

Thus, as many other papers in the area of digital humanities, we utilize topic modelling as a proxy to discourses. In particular, we apply the ``standard" Latent Dirichlet Allocation model~\cite[LDA]{blei2003latent} and its extension the Dynamic Topic Model~\cite[DTM]{blei2006dynamic}, which is developed specifically to tackle temporal dynamics in data. However, any model has its limitations and tends to exaggerate certain phenomena while missing other ones. We focus on the difference between models and try to reveal their limitations in historical data analysis from the point of view that is relevant for historical scholarship. 

Our main contributions are the following:
\begin{itemize}
\item We propose a \textbf{combined sampling, training and inference procedure} for applying topic models to large and imbalanced diachronic text collections.
\item We discuss differences between two topic models, paying special attention to how they \textbf{can be used to trace discourse dynamics}. 
\item We propose a method to quantify \textbf{topic prominence for a period} and thus to generalize document-wise topic assignment to a discourse level.
\item We \textbf{acknowledge and discuss the drawbacks of topic stretching,} which is typical for DTM. It is commonly known that DTM sometimes represents topics beyond the time period, but thus far there is no discussion in how researchers should tackle this for humanities questions.
\end{itemize}

In order to illustrate the appropriateness of the proposed methodology we discuss two use cases, one relating to discourses on church and religion and one that relates to education. The role of religion and education has been studied extensively in historical scholarship but there are no studies that deal with these topics through text mining of large-scale historical data. These two topics were chosen due to the the fact that the former was in general a discourse in decline relating to the process of secularization in Finnish society, whereas the latter increased in the second half of the nineteenth century and relates to the modernization of Finnish society and the inclusion of a larger share of the population in the sphere of basic education. In addition to these two interlinked discursive trends, we also use other examples to illustrate the strengths and weaknesses of LDA and DTM for this type of historical research.

\section{Data}
Our dataset is from the digitised newspaper collection of the National Library of Finland (NLF). This dataset contains articles from \textit{all} newspapers and most periodicals that have been published in Finland from 1771 to 1917. Several studies have used parts of this dataset to investigate such issues as the development of the public sphere in Finland, the evolution of ideological terms in nineteenth-century Finland and the changing vocabulary of Finnish newspapers \cite{vesanto2017applying,la_mela_finding_2019,kokko_suomenkielisen_2019,hengchen2019data,marjanen2019clustering,marjanen2019national,oiva_spreading_2020,salmi_reuse_2020,hengchen2020vocab}.

\begin{figure}[t]
\centering
\begin{subfigure}{.32\textwidth}
  \centering
  \includegraphics[width=\linewidth]{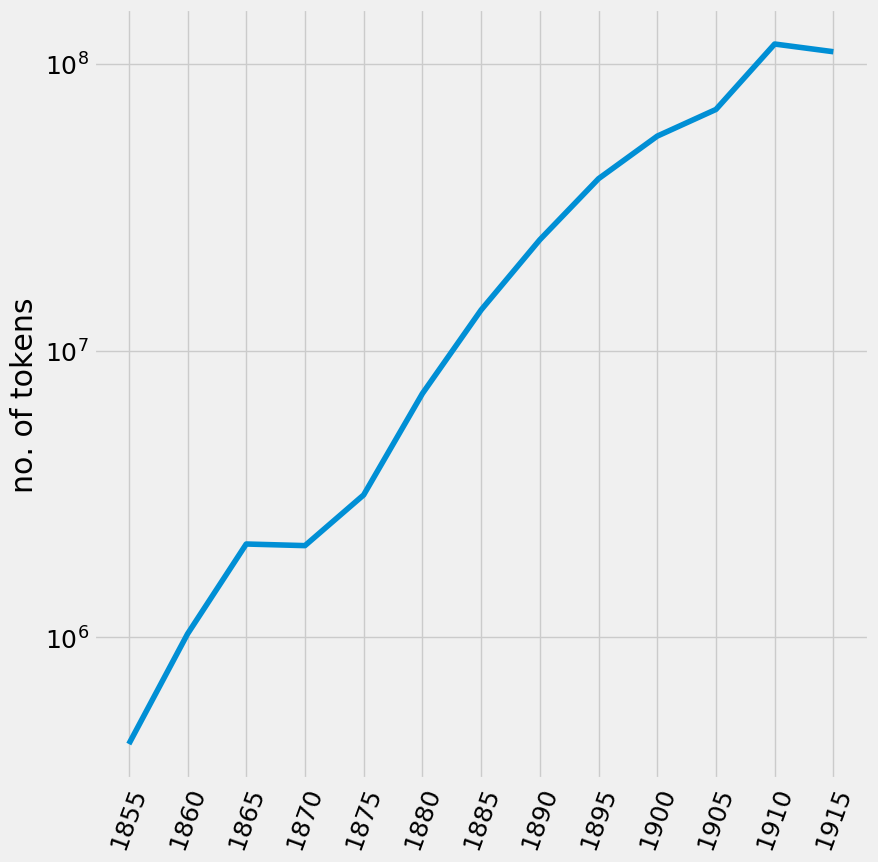}
  \caption{Corpus size }
  \label{fig:num_tokens}
\end{subfigure}
\begin{subfigure}{.32\textwidth}
  \centering
  \includegraphics[width=\linewidth]{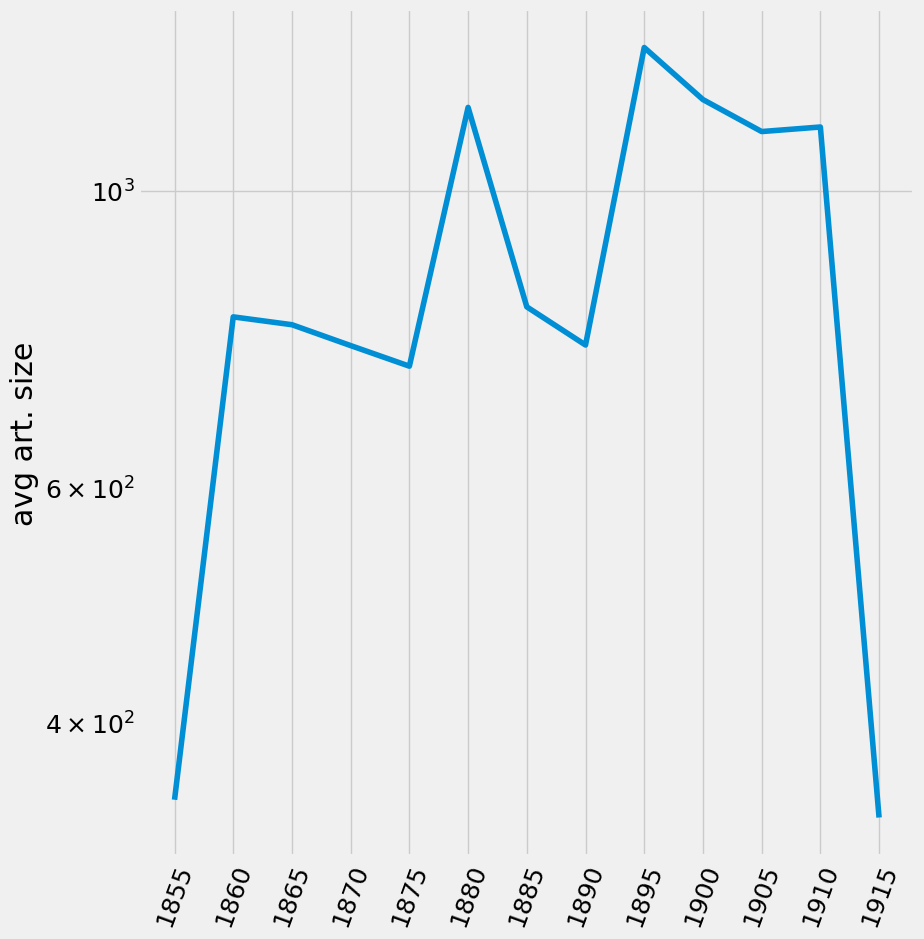}
  \caption{Average article size}
  \label{fig:article_length}
\end{subfigure}
\begin{subfigure}{.32\textwidth}
  \centering
  \includegraphics[width=\linewidth]{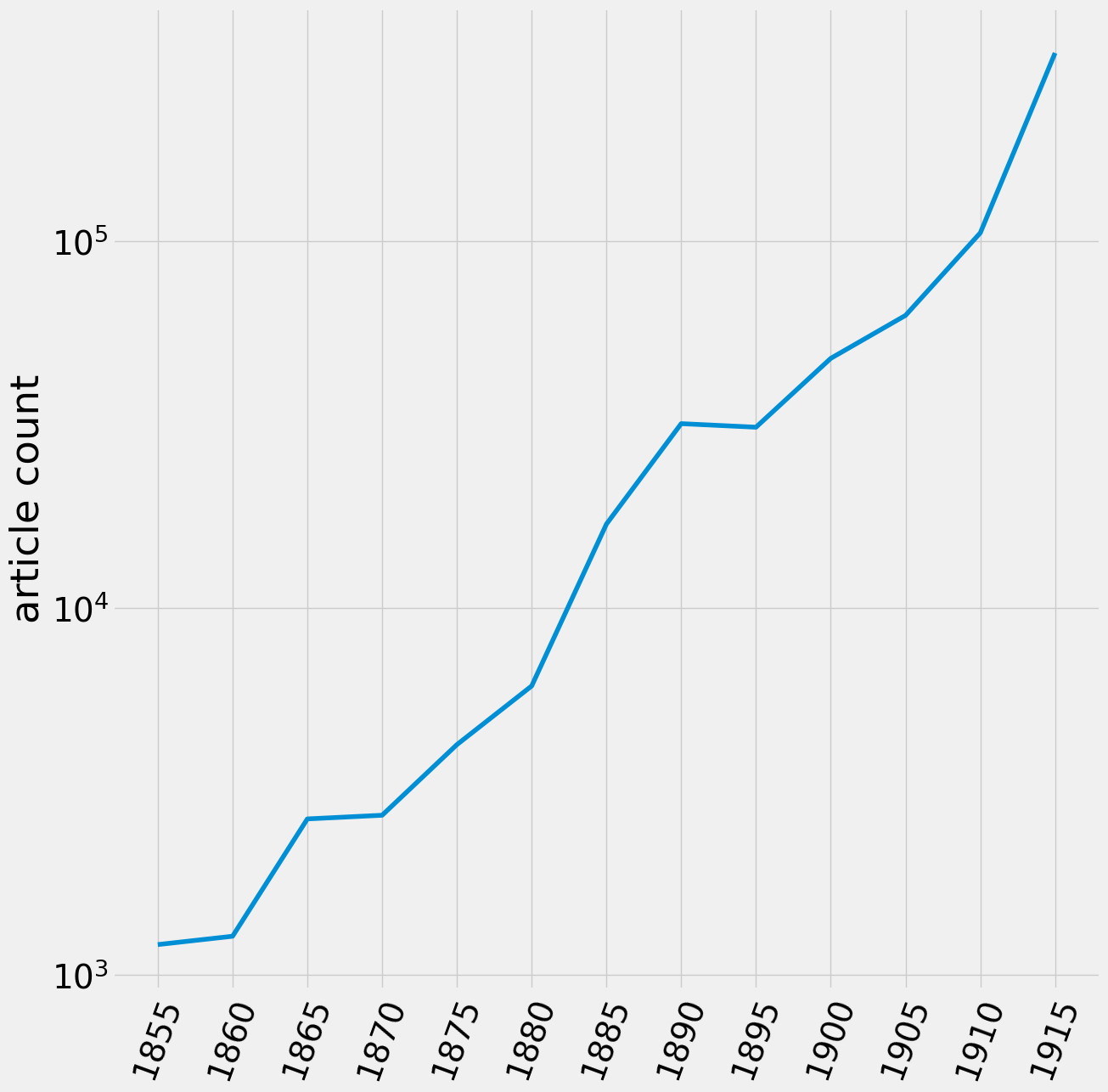}
  \caption{No. of articles}
  \label{fig:article_counts}
\end{subfigure}%
\caption{From left to right,  per year: overall corpus size of the NLF dataset, average article size, and article counts.}
\vspace*{-6mm}
\label{fig:nlf_data_numbers}
\end{figure}

The full collection includes articles in Finnish, Swedish, Russian, and German. In this work we focus only on the Finnish portion starting from 1854 because this is the point where we determined we have sufficient yearly data to train topic models. The resulting subset has over 3.6 million articles and is composed of over 2.2 billion tokens. Figure \ref{fig:num_tokens} shows that the number of tokens published per year in Finnish-language papers increased steadily. The average article has 526 tokens but article length varies widely from year to year, as seen in Figures~\ref{fig:article_length} and \ref{fig:article_counts} which show the average article length and the number of articles per year. As made clear by these figures, there is a noticeable difference in the number of articles and average article length after 1910. This shift does not reflect the actual articles in the  newspapers, but is the result of a change of OCR engine used to digitise the collection~\cite{21a9f51e784d453b8e7e050f66ffb265}. 

Still, even if the article segmentation differs in the latter period, Fig. \ref{fig:num_tokens} shows that there is steady increase in the vocabulary used in the Finnish-language newspapers published in the second half of the nineteenth century. They also covered more themes and regions. This entailed a process of diversification and modernization of the Finnish press, which has been widely discussed in historiography. As a collection, the newspapers vary a lot in style and focus. Some larger newspapers mainly contain political content, whereas others are rather specialised, and yet others thrived by giving a voice to the local public~\cite{tommila_suomen_1988,marjanen2019national,kokko_suomenkielisen_2019,sorvali_pyydan_2020}. This means that any analysis done on the entirety of the newspapers, like topic models, tend to balance out some of the differences between newspapers. This variety in the content, is also something that make newspapers such an interesting source material for historical research that is interesting in an overview of society. Although some issues were obviously not discussed because of taboo, courtesy or censorship, most of the themes present in public discourse are recorded in the newspapers and thus accessible to us in the present. Hence, we believe newspapers are an especially good source of assessing how the role of particular discourses changed over time.

\section{Topic Models}

\subsection{LDA}

Topic modelling is an unsupervised method to extract topics from a collection of documents. Typically, a topic is a probability-weighted list of words that together express a theme or idea of what the topic is about. One of the most popular topic modelling methods currently in use is Latent Dirichlet Allocation (LDA), which is ``a generative probabilistic model for collections of discrete data such as text corpora'' \cite{blei2003latent}. It has been extensively used in the digital humanities to extract certain themes from a collection of texts \cite{brauer2013historicizing}. In this model, a document is a mixture of topics and a topic is a probability distribution over a vocabulary. A limitation of LDA for historical research, in its vanilla form, is that it does not account for the temporal aspect of the data: every document in the collection is ``considered synchronic", as time is simply not a variable in the model. Many document collections such as news archives, however, are diachronic---the documents are from different points in time, and scholars wish to study the evolution of topics. 

There are different ways to overcome this limitation. One possibility is to split the data into time slices and train LDA separately on each slice. However, in this case LDA models for each slice would be independent of each other and there is no straightforward approach of matching topics from independent models trained on disjoint data. Another possibility, which we explore in this paper, is to train a single model for a subset of the whole data set over the entire time period and then use \emph{topic prominence} as proxy for the dynamics of discourses over time.

To do this, we compute the prominence of a topic in a given year by summing up the topic contribution for each document in that year and then normalise this number by the sum of all topic contributions from all topics for that year, as in Equation~\ref{eq:topic-prominence}.

\begin{equation}
\label{eq:topic-prominence}
P(z_k | y) = \frac{\sum_{j=1}^{|D_y|} P(z_k | d_j)}
{\sum_{i=1}^{T}\sum_{j=1}^{|D_y|} P(z_i | d_j)}
\end{equation}
where $y$ is a year in the dataset, $k$ is a topic index, $D_y$ is the number of documents in year $y$, $d_j$ is the $j^{th}$ document in year $y$ and $T$ is the number of topics in the model. 

The large size of the collection and its unbalanced nature is a problem for training topic models. It is computationally expensive to train a model with millions of articles and the resulting model would be heavily biased towards the latter years of newspaper collection because it has far more data. To overcome these issues, we sampled the collection such that we have a roughly similar data size for each year of the collection and as a result, we also get a vastly reduced dataset. However, to have a model of discourse dynamics that reflects the collection more closely, we compute topic prominence using the entire collection and not just the sampled portion. We do this by inferring the topic proportions of all the documents in the collection and using these inferred distributions to compute topic prominence.

\subsection{DTM}

As mentioned above, there are topic models that explicitly take into account the temporal dynamics of the data. One such model is the dynamic topic model (DTM). DTM is an extension of LDA that is designed to capture dynamic co-occurence patterns in diachronic data. In this model, the document collection is divided into discrete time slices and the model learns topics in each time slice with a contribution from the previous time slice. This results in topics that evolve slightly--words changing in saliency in relation to a topic--from one time step to the next. 

However, DTM also has its own limitations. It is based on an assumption that each topic should be to some extent present in each time slice, which is not always the case with real-world data such as news archives where events and themes can sometimes disappear and then re-appear at some point in the future. 

Perhaps more importantly for historical research, a weakness of DTM lies in its design: to accomplish alignment across time the topic model is fit across the whole vocabulary and thus smoothing between time slices is applied. As a result, events end up being ``spread out" before and after they are known to happen. This problem only becomes evident after a thorough analysis: similar models in different fields such as lexical semantic change present the same issue -- the dynamic topic model SCAN \cite{frermann-lapata-2016-bayesian} generates a ``plane" top word for the year 1700 (two centuries ahead of the Wright Flyer, and well before the word's first attested sense of ``aeroplane"), while similar model GASC \cite{perrone-etal-2019-gasc,mcgillivray2019computational} encounters the same weakness when modelling Ancient Greek. There is unfortunately no easy way to bypass this obstacle, which is particularly problematic when studying historical themes. 
\section{Related Work}
Topic models are widely used in the digital humanities and social sciences to draw insights from large-scale collections \cite{brauer2013historicizing} ranging from newspaper archives to academic journals. In this section, which we do not claim to be exhaustive, we discuss some of the previous works that aimed to capture historical trends in large data collections or used such collections to study discourses using topic models. All in all, these examples highlight that there is a need to discuss how topic models can be used to capture discursive change.

In \cite{newman2006probabilistic} the authors use Latent Semantic Analysis, another topic modelling method, to study historical trends in eighteenth-century colonial America with articles from the \emph{Pennsylvania Gazette}. Their work also used topic prominence to show, for instance, an increased interest in political issues as the country was heading towards revolution. The authors of \cite{yang2011topic} fit several topic models on Texan newspapers from 1829 to 2008. To discover interesting historical trends, the authors slice their data into four time bins, each corresponding to historically relevant periods. Such a slicing is also carried out in \cite{hengchen2017does}, where the author fits LDA models on Dutch-language Belgian socialist newspapers for three time slices that are historically relevant to the evolution of workers rights, with the aim of generating candidates for lexical semantic change.

Topic modelling has also been used in discourse analysis of newspaper data. In \cite{viola2019mining} the authors applied LDA to a selection of Italian ethnic newspapers published in the United States from 1898 to 1920 to examine the changing discourse around the Italian immigrant community, as told by the immigrants themselves, over time. They proposed a methodology combining topic modelling with close reading called discourse-driven topic modelling (DDTM). Another study examined anti-modern discourse in Europe from a collection of French-language newspapers \cite{bunout2020grasping}. In this case, however, the authors primarily use LDA as a tool to construct a sub-corpus of relevant articles that was then used for further analysis. Modernization was also an issue in the study of Indukaev \cite{indukaev_studying_2021}, who uses LDA and word embeddings to study changing ideas of technology and modernization in Russian newspapers during the Medvedev and Putin presidencies. 

LDA was not designed for capturing trends in diachronic data and so several methods have been developed to address this, such as DTM, Topics over Time \cite[TOT]{wang2006topics}, and the more recent Dynamic Embedded Topic Model \cite[DETM]{dieng2019dynamic}, an extension of DTM that incorporates information from word embeddings during training. As far as we are aware, DTM and TOT have not been used for historical discourse analysis or applied to large-scale data collections. In the original papers presenting these methods, DTM was applied to 30,000 articles from the journal \textit{Science} covering 120 years and TOT was applied to 208 State of the Union Presidential addresses covering more than 200 years. This was to demonstrate the evolution of scientific trends for the former and the localisation of significant historical events for the latter. Recently DETM was applied on a dataset of modern news articles about the COVID-19 pandemic where the authors observed differences between countries in how the pandemic and the reactions to it were framed \cite{li2020global}.

In the mentioned cases researchers tackle the interpretative part of using topic models for humanistic research in different ways. Like Pääkkönen and Ylikoski \cite{paakkonen_humanistic_2020} state, they toggle between some sort of topic realism, that is, using topic models to grasp something that exists in the data, and topic instrumentalism, that is, using topic models to find something that can be further studied. Only Bunout \cite{bunout2020grasping} is a clear case of topic instrumentalism. All the other studies depart from some sort of realist position, and attempt to grasp policy shifts, ideas, discourses or framings of topics through topic models, but end up with correctives of some kind by highlighting the interpretative element \cite{newman2006probabilistic,viola2019mining}, by deploying formal evaluation by historians \cite{hengchen2017does} or by using other quantitative methods to fine tune the results \cite{indukaev_studying_2021}. The interpretative aspect seems especially important when it comes to deciding on what researchers use the topics study as they can reasonable relate to historical discourses, the semantics of related words, or simply ideas. How the topics are seen to represent these or, more likely, how the researchers use the topics to make an interpretation about these based on the topics, requires an element of interpretation\cite{paakkonen_humanistic_2020}. 

\section{Use cases}
What a discourse is, has been heavily theorised within the different strands of discourse analysis~\cite{angermuller_discourse_2014}, but the advent of digital methods that can handle large textual data sets require quite some adjustment of discourse analysis as we know it. Like this article, others have turned to topic models to grasp changes in discourse \cite{viola2019mining,bunout2020grasping}, but this article seeks specifically to discuss the interpretation that is required when we use topic models to study discourse dynamics. The probabilistic topic models set clear boundaries between topics and in doing so might merge or separate things that historians might regard as coherent topics. However, where the probabilistic model enforces boundaries, human interpretation in general is very bad at setting those boundaries and usually just identifies the core of a discourse or topic, but cannot say where it ends.

To get at the tension between topics and discourses, we approached the material without a predefined idea about which topics we wanted to study in order to keep the study as data-driven as possible. Our interest was to use topic modelling to capture topics that could in a meaningful way be related to societal discourses, that is themes that cannot be narrowed down to individual words, but still are reasonably coherent and form at least loose topics. To this end, we trained topic models with 
$k \in \{30;50\}$, inferred topic distributions
for the whole collection and inspected models by carefully going through the top words in each topic and using PyLDAVis\footnote{\url{https://github.com/bmabey/pyLDAvis}} \cite{sievert2014ldavis} to study overlap between topics and salience of terms per topic in LDA and heatmap visualizations for DTM. All topics were annotated and evaluated from the point of view of historical interpretation. We then opted to use the 50-topic model to study discourse changes over time. As is common, a portion of the topics seemed incoherent or were clearly the result of the layout in newspapers (e.g. boilerplate articles about prices etc.) and did not produce interesting information about societal discourses. Further, some of the topics clearly overlap, so that a cluster of 2-5 topics can reasonably be seen as related to a particular societal discourse. The advantage of choosing fifty topics over thirty lies precisely in the possibility of merging topics later on in interpretation, while splitting them is more difficult.

To discuss the benefits of LDA and DTM, we chose to focus on two specific themes, the discourse relating to religion and religious offices, and education. They are both rather neatly identifiable in the data, but display different trends. The former is in decline over the period of interest, whereas the latter increases in topic prominence. They can also be related to large scale processes in Finland, religious discourse to the secularization of society and education to the modernization of civic engagement.

\subsection{DTM and stretching of topics}

The two topic modelling methods perform in somewhat different ways. As mentioned, DTM is designed to incorporate temporal change in the topics, which means it includes a stronger sense of continuity in its representations of data. Whether or not this is desirable, depends on the research question, but our contention is that for studies interested in discursive change, this is either a problem or at least it is something that needs to be factored in making the historical interpretation. If we want to understand when certain discourses became dominant, declined, or even disappeared, this type of stretching cannot be allowed.  

\begin{figure}[t]
\centering
\begin{subfigure}{.45\columnwidth}
    \centering
    \includegraphics[width=0.9\columnwidth]{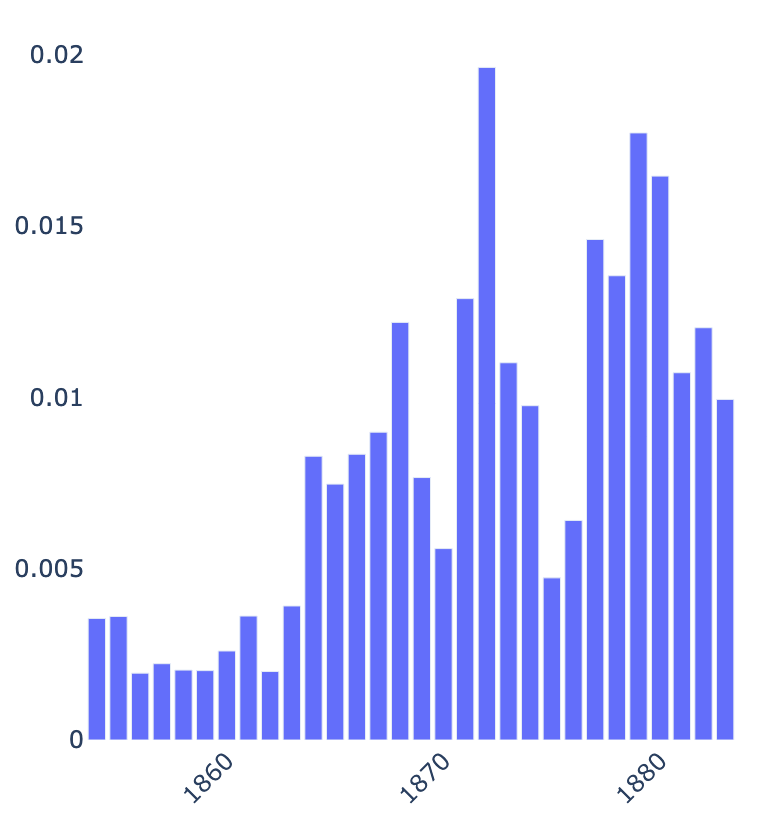}
    \caption{Introduction of the Finnish mark in 1860}
    \label{fig:markka_dtm}
\end{subfigure}
\begin{subfigure}{.45\columnwidth}
    \vspace*{7mm}
    \centering
    \includegraphics[width=0.9\columnwidth]{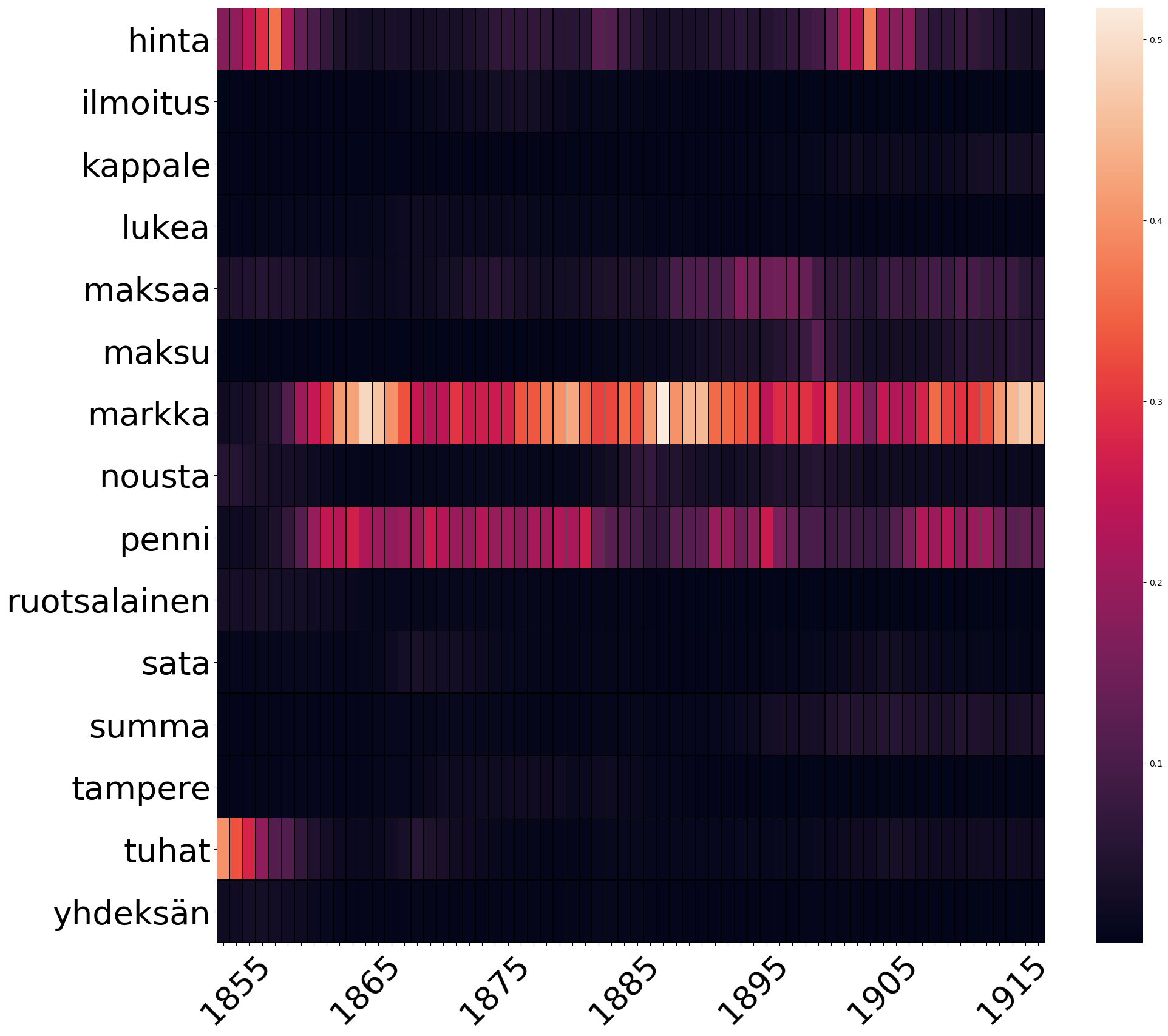}
    \caption{Heat map of terms linked to the introduction of the Finnish mark in 1860} 
    \label{fig:markka_heatmap_dtm}
\end{subfigure}
\caption{Topic related to the introduction of the Finnish mark in 1860 (DTM)}
\vspace*{-3mm}
\label{fig:markka_dtm_all}
\end{figure}

An exceptionally illustrative example of stretching among our fifty topics, is an introduction of the Finnish mark as a currency (Fig. \ref{fig:markka_dtm}). With top words such as ``mark", ``penny", ``price", ``thousand", ``pay" etc. the topic comes across as one with high internal coherence.\footnote{To make the reading easier for people unfamiliar with the Finnish language, we use English translations in the text. ``Mark" = \textit{markka}, ``penny" = \textit{penni}, ``price" = \textit{hinta}, ``thousand" = \textit{tuhat}, ``pay" = \textit{maksu} and \textit{maksaa}.} We also see that the topic grows in prominence over time, from being relatively modest in the 1850s to gradually increased prominence after 1860. This makes sense, as the mark was adopted as currency in the year 1860 and after that self-evidently figured in public discourse. However, when we look at a heatmap visualization of the topic (Fig. \ref{fig:markka_heatmap_dtm}), we see how the topic stretches from the period 1854--1859 to the period 1860--1917, that is, from the period before the introduction of the mark to the period it was in use. After 1860 the words ``mark" and ``penny" are by far the most dominant terms in the topic, but for the period before 1860, the dominant terms are ``price" and ``thousand." It is clear that ``mark'', ``penny'', ``price'', and ``thousand'' are words that can belong to the same topic, but the heatmap representation clearly shows that the focus in the topic shifts. It is almost as if two related topics are merged as to represent one topic over the whole time period. In a situation where a historical interpretation highlights a change in past discourse, DTM produces continuity.

While there is obviously no right answer as to when one topic is stretched a bit or when different topics are simply merged together to provide a temporally continuous topic, it seems that DTM is especially problematic if one wants to study discourses that emerge or disappear in the middle of a time period studied. This means that any historical analysis using DTM requires a component of historical interpretation of not only topic coherence, but also topic coherence \textit{over time}. Here, relying on word embeddings like in \cite{indukaev_studying_2021} can help, but this is primarily a task for evaluating the topics.

The speed of topic evolution can be controlled by a parameter in the DTM model. However, the `ideal' amount of stretching is difficult to assess. For analysing discourse, this might in some cases be productive as it can point at links between nearby discourses, but is largely problematic as it hides discontinuities in the data. It becomes even problematic when dealing with material factors, like the introduction of the Finnish mark, as the stretching effect is likely to produce anachronistic representations, that is, placing something in the wrong period of time. Dealing with anachronism can perhaps be seen as one of the cornerstones of the historian's profession, which makes DTM as an anachronism prone method a poor match for historical study. Avoiding anachronisms completely is impossible, most historians would agree, but knowing when to avoid them and how to communicate about anachronistic elements in historical interpretation is key to history as a discipline \cite{syrjamaki_sins_2011}.

\subsection{Religion and secularization}

\begin{figure}[t!]
\centering
\begin{subfigure}{.32\textwidth}
    \centering
    \includegraphics[width=\columnwidth]{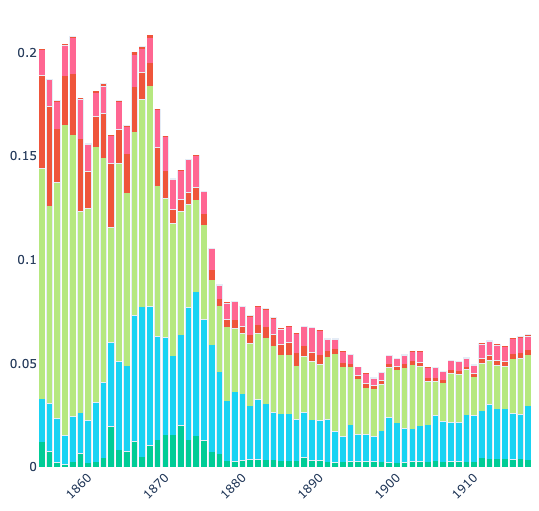}
    \caption{Topics related to religion on decline (LDA)}
    \label{fig:religion_lda}
\end{subfigure}
\begin{subfigure}{.32\textwidth}
    \centering
    \vspace*{2mm}
    \includegraphics[width=\columnwidth]{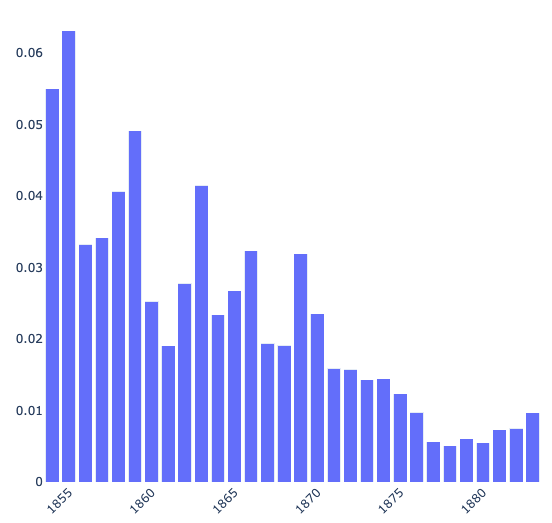}
    \caption{Development of religious topic (chaplain, priest and office) over time}
    \label{fig:religion_dtm}
\end{subfigure}
\begin{subfigure}{.32\textwidth}
    \centering
    \includegraphics[width=\columnwidth]{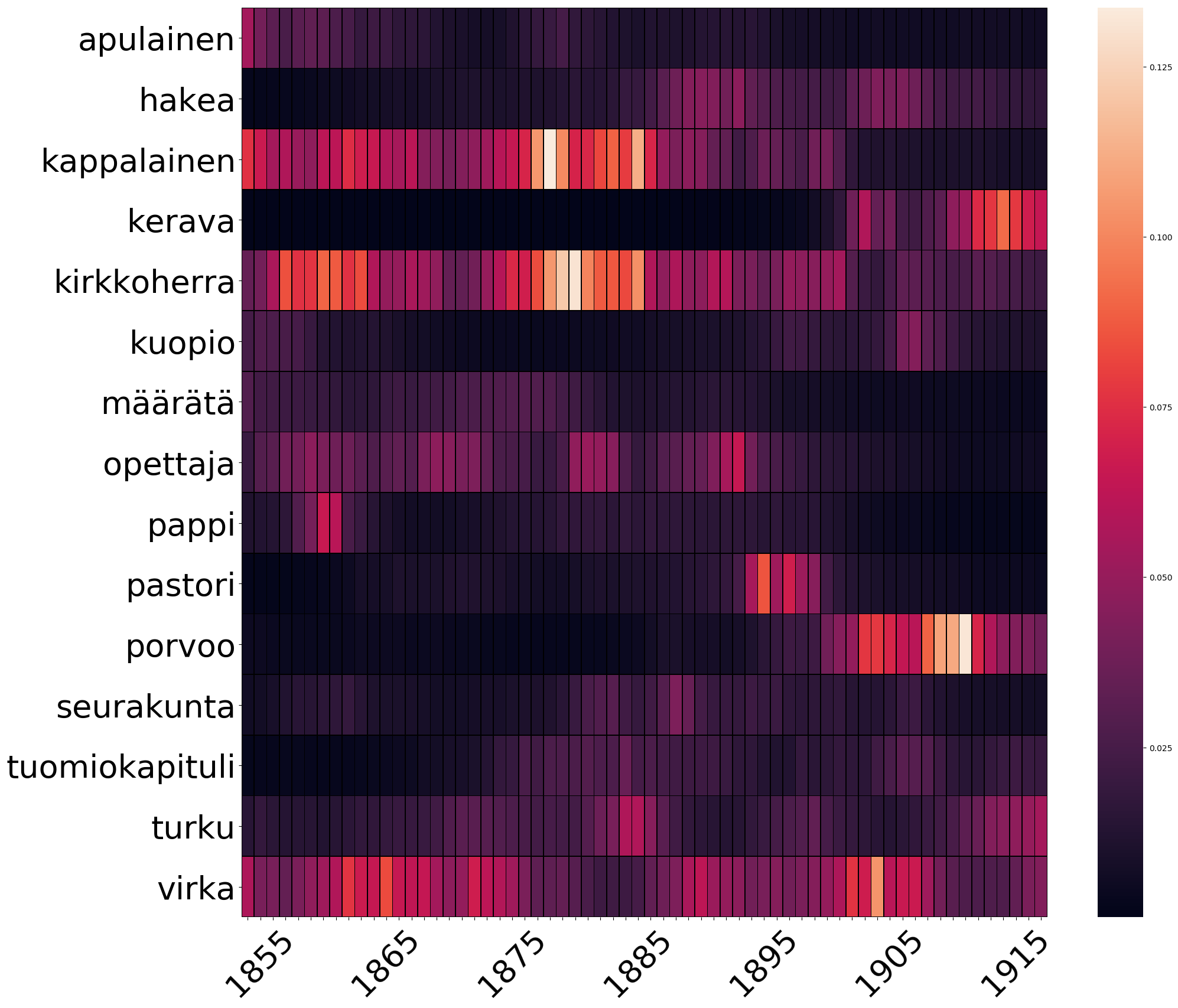}
    \caption{Heatmap of terms linked to office of religion topic}
    \label{fig:heatmap_religion}
\end{subfigure}
    \caption{Religious topics in LDA (a) and DTM (b,c)}
\vspace*{-3mm}
\label{fig:religion_all}
\end{figure}

Our model performed well in grasping topics that relate to religion. The initial expectation regarding the discourse dynamics was that religious topics would be in decline. We hoped that using a topic model would be a way of showing this quantitatively. Results obtained from both LDA and DTM, presented in Figures \ref{fig:religion_lda} and \ref{fig:religion_dtm} respectively, harmonize with our initial hypothesis, but do so differently. The DTM and LDA outputs cannot be aligned in any other way than manual interpretation by domain experts. In order to inspect the discourse dynamics of religious topics, we have combined several topics that related to religious theme in the LDA model, whereas in the latter, DTM model, we only chose one topic to be represented. 

To our knowledge, topic models have not been used to study discursive change regarding secularization. However, in line with some earlier qualitative assessments \cite{juva_valtiokirkosta_1960}, we hypothesize that this decline in religious discourse entails two interrelated developments: 1) Religion did not disappear from public discourse, but instead changed and disappeared from certain \textit{types} of discourses. In the early nineteenth century, religion had a much more holistic presence in public discourse, meaning that religious metaphors and religious expressions and topics were used at a much vaster scale. 2) Over the course of the nineteenth century, religious topics became more focused. This means a segmentation of public discourse so that religious topics were increasingly confined to particular journals or genres.

Keeping in mind the issue of stretching with DTM, we can look into the shifting saliency of words within the topic of religious offices and notice a shifting focus over time (Fig. \ref{fig:heatmap_religion}). In the early 1900s terms relating to "holding an office" and names of particular congregations become more dominant in the topic. This, again, suggests that DTM as a method does some stretching. There is a downside and an upside to this. On the one hand, the stretching distorts the topic prominence a bit by making it look like there is more continuity than in the LDA visualization. However, this may not be that crucial as the declining trends in Fig. \ref{fig:religion_lda} and Fig. \ref{fig:religion_dtm} are rather similar. On the other hand, the stretching may be good for detecting conceptual links between different groups of words. In this particular case the stronger link between religious offices and some towns like Kerava and Porvoo, is probably indicative of a move of religious discourse from an overarching question to something that is more likely dealt with in conjunction to matters at local parishes. That is, religious offices were more often than before dealt with in connection to local congregations. This is in line with our above-mentioned assumption about religious discourse becoming more distinct.

\subsection{Education and modernity}

While we expected religious themes to decline and become less central, we assumed there would be some themes that partly overlap with religion, but also would show an increasing trend. One example of this is the topic of education, which has historically been heavily interwoven with the church, but at the same time when basic education became available for a higher amount of people, it also became central in questioning the role of the church and religion. Education in  nineteenth-century Finland was both central for ensuring conformity of the Lutheran faith, but paradoxically also was a vehicle of secularization. \cite{hanska_huoneentaulun_nodate}

\begin{figure}
\centering
\begin{subfigure}{.43\textwidth}
    \centering
    \includegraphics[width=0.9\columnwidth]{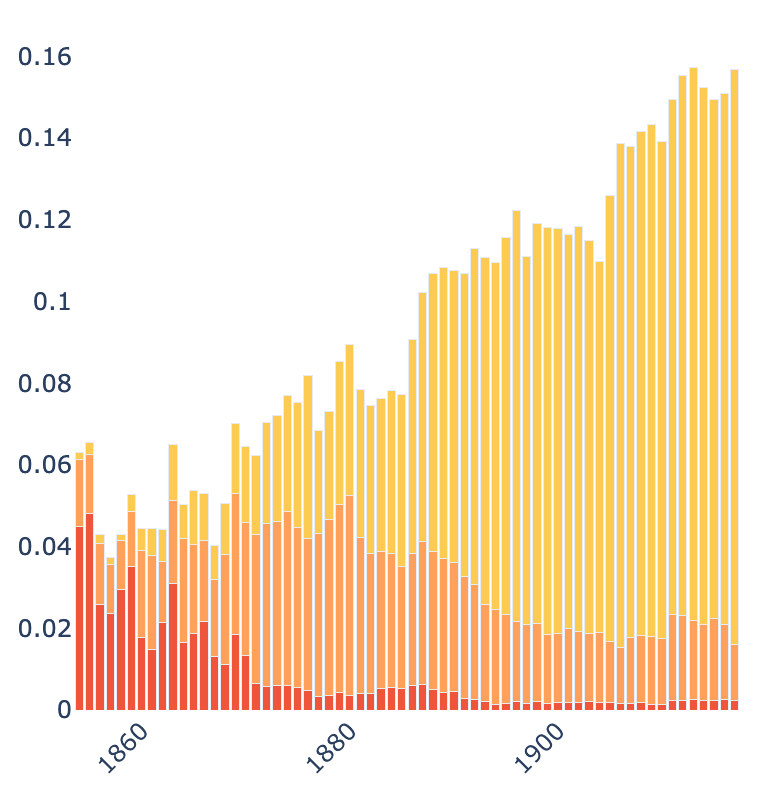}
    \caption{Development of education topic over time (LDA)}
    \label{fig:education_lda}
\end{subfigure}
\begin{subfigure}{.43\textwidth}
    \centering
    \includegraphics[width=0.9\columnwidth]{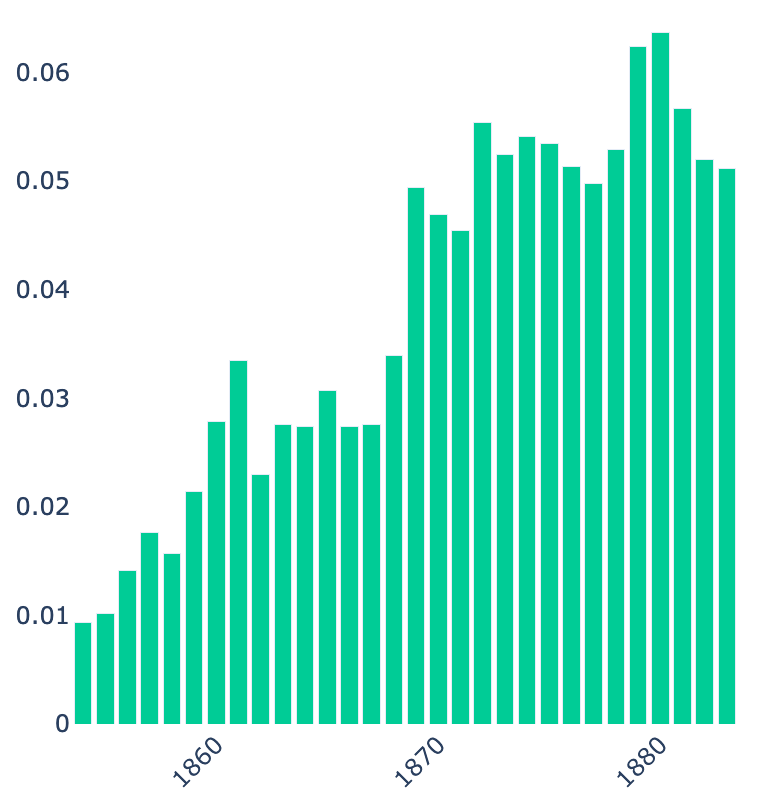}
    \caption{Development of education topic over time (DTM)}
    \label{fig:education_dtm}
\end{subfigure}
    \caption{Education topic in LDA and DTM}
\vspace*{-1mm}
\label{fig:education_all}
\end{figure}

As in the case of religious discourse, alignment between DTM and LDA can only be made through human interpretation. It seems, that in this case DTM captures one topic that is fairly coherent, revolves around education and schooling, and is on the rise in the research period (Fig. \ref{fig:education_dtm}). For LDA, this is not the case, as an PyLDAVis inspection of most salient words across all fifty topics show that words like ``school" and ``folk school" appear mostly in three topics of which two are in decline and one heavily on the rise (Fig. \ref{fig:education_lda}). 

Interestingly, LDA and DTM seem to be pointing at a similar historical development. The two declining LDA topics are based on their most salient terms and are more focused on schools as buildings and institutions as well as teaching as a profession, whereas the topic on the rise includes salient vocabulary relating to, not only schools, but also meetings, civic engagements, and decision making. The DTM topic at hand shows a similar development which can be inspected in a heatmap of most salient terms over time. The terms ``school", ``child", and ``teacher" dominate early in the period. By the end of the period the topic becomes broader, and terms like ``municipality" and ``meeting" have become more salient than the vocabulary relating to schools. Here the stretching of DTM creates the links that are also visible in the three LDA topics, and it shows a transformation in which educational issues are present in the whole topic, but focus shifts from concrete schools to civic engagement.

\section{Conclusions}

Our focus in this text has been on discourses that cannot be reduced to mere words, isolated events or particular people, but concern broader societal topics that either declined or gained in prominence. The interpretation of these topics and their contextualisation to nineteenth-century Finnish newspapers revealed clear topical cores that can be interpreted as an encouraging point of departure for further explorations based on topic models when aiming to understand Finnish public discourse through historical newspapers. 

In this paper, we have learned that although it is difficult to pinpoint exactly where a discourse or topic ends, LDA and DTM can fairly reliably grasp many semi-coherent themes in past discourse and help us study the dynamics of discourses. However, our comparison of LDA and DTM as methods for getting at past discourse also shows that both methods require a very strong interpretative element in analysing historical discourses. DTM is much more prone to stretch or even merge topics, which requires an interpretative assessment of whether the stretching highlights interesting historical continuities or if it hides historical discontinuities that would require attention. We found that producing heatmaps of term saliency over time for each topic is a very useful way of doing this type of assessment. For LDA, stretching is not so much a problem, but often it seems interpretation is needed in seeing which topics logically relate to one another. While historical discourse analysis is traditionally tied strongly to a tradition of hermeneutic interpretation, the use of topic models to grasp discourse dynamics does not remove that need even if they allow for a quantification of discourse dynamics over time. 

While we regard stretching in DTM as a predominantly negative feature, in some cases it can be useful. In the topics relating to education discussed above, the stretching in DTM actually points out links in discourses and is quite productive for the interpretative process of trying to figure out discourse dynamics. However, also in this case, the relevance of historical interpretation should be highlighted because it is very hard to tell whether the stretching of topics is an accurate reflection of the data or a shortcoming of the model. This can be addressed only by relating visualisations of topics to existing historical research and reading source texts. Humanities scholars are in general very good at making such interpretations, but it also needs to be noted that when we move further into the domain interpretative scholarship, we also lose some of the benefits of working with quantifying models. While it would be foolish to claim that a topic model represents data in a way that it provides simple facts about historical development, our use cases show that if we seek to find more reliable quantification LDA may provide better results than DTM. Further, using LDA moves the interpretative stage further down in the research process, as it is likely to be about evaluating the connections between different topics over time. In DTM, the interpretation is likely moved forward to an evaluation of how well the algorithm did this merging topics. On this sense, our take on topic models harmonises with \cite{paakkonen_humanistic_2020} who stress the role of humanistic interpretation, but for the sake of transparency suggest pushing the interpretation stage later in the research process.

\section*{Acknowledgements}
This work has been supported by the European Union Horizon 2020 research and innovation programme under grant 770299 (NewsEye). SH is funded by the project \textit{Towards Computational Lexical Semantic Change Detection} supported  by the Swedish Research Council (2019--2022; dnr 2018-01184).

\bibliographystyle{splncs04}
\bibliography{biblio}

\begin{thebibliography}{10}
\providecommand{\url}[1]{\texttt{#1}}
\providecommand{\urlprefix}{URL }
\providecommand{\doi}[1]{https://doi.org/#1}

\bibitem{angermuller_discourse_2014}
Angermuller, J., Maingueneau, D., Wodak, R. (eds.): The discourse studies
  reader: {Main} currents in theory and analysis. John Benjamins Publishing,
  Amsterdam, the Netherlands ; Philadelphia PA (2014)

\bibitem{blei2006dynamic}
Blei, D.M., Lafferty, J.D.: Dynamic topic models. In: Proceedings of the 23rd
  international conference on Machine Learning. pp. 113--120 (2006)

\bibitem{blei2003latent}
Blei, D.M., Ng, A.Y., Jordan, M.I.: Latent {D}irichlet allocation. Journal of
  Machine Learning Research  \textbf{3}(Jan),  993--1022 (2003)

\bibitem{brauer2013historicizing}
Brauer, R., Fridlund, M.: Historicizing topic models, a distant reading of
  topic modeling texts within historical studies. In: International Conference
  on Cultural Research in the context of Digital Humanities, St. Petersburg:
  Russian State Herzen University (2013)

\bibitem{bunout2020grasping}
Bunout, E.: Grasping the anti-modern discourse on {E}urope in the digitised
  press or can text mining help identify an ambiguous discourse?  (2020)

\bibitem{dieng2019dynamic}
Dieng, A.B., Ruiz, F.J., Blei, D.M.: The dynamic embedded topic model. arXiv
  preprint arXiv:1907.05545  (2019)

\bibitem{frermann-lapata-2016-bayesian}
Frermann, L., Lapata, M.: A {B}ayesian model of diachronic meaning change.
  Transactions of the Association for Computational Linguistics  \textbf{4},
  31--45 (2016)

\bibitem{hanska_huoneentaulun_nodate}
Hanska, J., Vainio-Korhonen, K. (eds.): Huoneentaulun maailma: kasvatus ja
  koulutus {Suomessa} keskiajalta 1860-luvulle. Suomalaisen {Kirjallisuuden}
  {Seuran} toimituksia, 1266:1, Suomalaisen kirjallisuuden seura, Helsinki
  (2010), publication Title: Huoneentaulun maailma : kasvatus ja koulutus
  Suomessa keskiajalta 1860-luvulle

\bibitem{hengchen2017does}
Hengchen, S.: When Does it Mean? Detecting Semantic Change in Historical Texts.
  Ph.D. thesis, Universit{\'e} libre de Bruxelles (2017)

\bibitem{hengchen2019data}
Hengchen, S., Ros, R., Marjanen, J.: A data-driven approach to the changing
  vocabulary of the nation in {English, Dutch, Swedish and Finnish} newspapers,
  1750-1950. In: Proceedings of the Digital Humanities (DH) conference (2019)

\bibitem{hengchen2020vocab}
Hengchen, S., Ros, R., Marjanen, J., Tolonen, M.: A data-driven approach to
  studying changing vocabularies in historical newspaper collections. Digital
  Scholarship in the Humanities  (2020)

\bibitem{indukaev_studying_2021}
Indukaev, A.: Studying {Ideational} {Change} in {Russian} {Politics} with
  {Topic} {Models} and {Word} {Embeddings}. In: Gritsenko, D., Wijermars, M.,
  Kopotev, M. (eds.) Palgrave {Handbook} of {Digital} {Russia} {Studies}.
  Palgrave Macmillan, Basingstoke (2021)

\bibitem{juva_valtiokirkosta_1960}
Juva, M.: Valtiokirkosta kansankirkoksi: {Suomen} kirkon vastaus
  kahdeksankymmentäluvun haasteeseen. WSOY, Porvoo (1960)

\bibitem{kokko_suomenkielisen_2019}
Kokko, H.: Suomenkielisen julkisuuden nousu 1850-luvulla ja sen
  yhteiskunnallinen merkitys. Historiallinen Aikakauskirja  \textbf{117}(1),
  5--21 (2019)

\bibitem{la_mela_finding_2019}
La~Mela, M., Tamper, M., Kettunen, K.: Finding {Nineteenth}-century {Berry}
  {Spots}: {Recognizing} and {Linking} {Place} {Names} in a {Historical}
  {Newspaper} {Berry}-picking {Corpus}. In: Navarretta, C., Agirrezabal, M.,
  Maegaard, B. (eds.) {DHN} 2019 - {Digital} {Humanities} in the {Nordic}
  {Countries}. pp. 295--307. {CEUR} {Workshop} {Proceedings}, CEUR (2019),
  \url{https://cst.dk/DHN2019/DHN2019.html}

\bibitem{li2020global}
Li, Y., Nair, P., Wen, Z., Chafi, I., Okhmatovskaia, A., Powell, G., Shen, Y.,
  Buckeridge, D.: Global surveillance of covid-19 by mining news media using a
  multi-source dynamic embedded topic model. In: Proceedings of the 11th ACM
  International Conference on Bioinformatics, Computational Biology and Health
  Informatics. pp. 1--14 (2020)

\bibitem{21a9f51e784d453b8e7e050f66ffb265}
M{\"a}kel{\"a}, E., Tolonen, M., Marjanen, J., Kanner, A., Vaara, V., Lahti,
  L.: Interdisciplinary collaboration in studying newspaper materiality. In:
  {Krauwer }, S., Fišer, D. (eds.) Twin Talks Workshop at DHN 2019. pp.
  55--66. CEUR Workshop Proceedings, CEUR-WS.org, Germany (2019)

\bibitem{marjanen2019clustering}
Marjanen, J., Pivovarova, L., Zosa, E., Kurunm{\"a}ki, J.: Clustering
  ideological terms in historical newspaper data with diachronic word
  embeddings. In: 5th International Workshop on Computational History,
  HistoInformatics 2019. CEUR-WS (2019)

\bibitem{marjanen2019national}
Marjanen, J., Vaara, V., Kanner, A., Roivainen, H., M{\"a}kel{\"a}, E., Lahti,
  L., Tolonen, M.: A national public sphere? {A}nalyzing the language,
  location, and form of newspapers in {F}inland, 1771--1917. Journal of
  European Periodical Studies  \textbf{4}(1),  54--77 (2019)

\bibitem{mcgillivray2019computational}
McGillivray, B., Hengchen, S., L{\"a}hteenoja, V., Palma, M., Vatri, A.: A
  computational approach to lexical polysemy in {Ancient Greek}. Digital
  Scholarship in the Humanities  \textbf{34}(4),  893--907 (2019)

\bibitem{newman2006probabilistic}
Newman, D.J., Block, S.: Probabilistic topic decomposition of an
  eighteenth-century american newspaper. Journal of the American Society for
  Information Science and Technology  \textbf{57}(6),  753--767 (2006)

\bibitem{oiva_spreading_2020}
Oiva, M., Nivala, A., Salmi, H., Latva, O., Jalava, M., Keck, J., Domínguez,
  L.M., Parker, J.: Spreading {News} in 1904: {The} {Media} {Coverage} of
  {Nikolay} {Bobrikov}’s {Shooting}. Media History  \textbf{26}(4),  391--407
  (Oct 2020). \doi{10.1080/13688804.2019.1652090},
  \url{https://www.tandfonline.com/doi/full/10.1080/13688804.2019.1652090}

\bibitem{perrone-etal-2019-gasc}
Perrone, V., Palma, M., Hengchen, S., Vatri, A., Smith, J.Q., McGillivray, B.:
  {GASC}: Genre-aware semantic change for {Ancient Greek}. In: Proceedings of
  the 1st International Workshop on Computational Approaches to Historical
  Language Change. pp. 56--66. Association for Computational Linguistics,
  Florence, Italy (Aug 2019). \doi{10.18653/v1/W19-4707},
  \url{https://www.aclweb.org/anthology/W19-4707}

\bibitem{paakkonen_humanistic_2020}
Pääkkönen, J., Ylikoski, P.: Humanistic interpretation and machine learning.
  Synthese  (Sep 2020). \doi{10.1007/s11229-020-02806-w},
  \url{http://link.springer.com/10.1007/s11229-020-02806-w}

\bibitem{salmi_reuse_2020}
Salmi, H., Paju, P., Rantala, H., Nivala, A., Vesanto, A., Ginter, F.: The
  reuse of texts in {Finnish} newspapers and journals, 1771–1920: {A} digital
  humanities perspective. Historical Methods: A Journal of Quantitative and
  Interdisciplinary History pp. 1--15 (Sep 2020).
  \doi{10.1080/01615440.2020.1803166},
  \url{https://www.tandfonline.com/doi/full/10.1080/01615440.2020.1803166}

\bibitem{sievert2014ldavis}
Sievert, C., Shirley, K.: Ldavis: A method for visualizing and interpreting
  topics. In: Proceedings of the workshop on interactive language learning,
  visualization, and interfaces. pp. 63--70 (2014)

\bibitem{sorvali_pyydan_2020}
Sorvali, S.: "{Pyydän} nöyrimmästi sijaa seuraavalle" – {Yleisönosaston}
  synty, vakiintuminen ja merkitys autonomian ajan {Suomen} lehdistössä.
  Historiallinen Aikakauskirja  \textbf{118}(3),  324--339 (2020)

\bibitem{syrjamaki_sins_2011}
Syrjämäki, S.: Sins of a historian: {Perspectives} on the problem of
  anachronism. Ph.D. thesis, Tampere University Press, Tampere (2011), oCLC:
  816367378

\bibitem{tommila_suomen_1988}
Tommila, P., Landgrén, L.F., Leino-Kaukiainen, P.: Suomen lehdistön historia
  1. {Sanomalehdistön} vaiheet vuoteen 1905. Kustannuskiila, Kuopio (1988)

\bibitem{vesanto2017applying}
Vesanto, A., Nivala, A., Rantala, H., Salakoski, T., Salmi, H., Ginter, F.:
  Applying {BLAST} to text reuse detection in finnish newspapers and journals,
  1771-1910. In: Proceedings of the NoDaLiDa 2017 Workshop on Processing
  Historical Language. pp. 54--58 (2017)

\bibitem{viola2019mining}
Viola, L., Verheul, J.: {Mining ethnicity: Discourse-driven topic modelling of
  immigrant discourses in the USA, 1898--1920}. Digital Scholarship in the
  Humanities  (2019)

\bibitem{wang2006topics}
Wang, X., McCallum, A.: Topics over time: a non-markov continuous-time model of
  topical trends. In: Proceedings of the 12th ACM SIGKDD international
  conference on Knowledge discovery and data mining. pp. 424--433 (2006)

\bibitem{yang2011topic}
Yang, T.I., Torget, A., Mihalcea, R.: Topic modeling on historical newspapers.
  In: Proceedings of the 5th ACL-HLT Workshop on Language Technology for
  Cultural Heritage, Social Sciences, and Humanities. pp. 96--104 (2011)

\end{thebibliography}

\end{document}